\documentclass[letterpaper, 10 pt, journal, twoside]{IEEEtran}

\IEEEoverridecommandlockouts                              

\usepackage{graphics} 
\usepackage{epsfig} 
\usepackage{times} 
\usepackage{amsmath} 
\usepackage{amssymb}  
\usepackage{amsthm}
\usepackage{listings}
\usepackage{paralist}
\usepackage{algorithm}
\usepackage{algorithmicx}
\usepackage{algpseudocode}
\usepackage{color}
\usepackage{flushend}
\usepackage{hyperref}

\definecolor{myblue}{rgb}{0.0,0.2,0.9}

\newtheorem{definition}{Definition}
\newtheorem{theorem}{Theorem}

\title{
\textsc{Sablas}: Learning Safe Control for Black-box Dynamical Systems
}

\begin{document}

\author{Zengyi Qin$^{1}$,  Dawei Sun$^{2}$, and Chuchu Fan$^{1}$%
\thanks{Manuscript received: September 9, 2021; Revised December 11, 2021; Accepted January 5, 2022.}
\thanks{This paper was recommended for publication by Editor Clement Gosselin upon evaluation of the Associate Editor and Reviewers' comments. This work was supported by the Defense Science and Technology Agency in Singapore, but this article solely reflects the opinions and conclusions of its authors and not DSTA Singapore or the Singapore Government.} 
\thanks{$^{1}$Zengyi Qin and Chuchu Fan are with Massachusetts Institute of Technology, Cambridge, MA, 02139 USA.
        {\tt\footnotesize \{qinzy, chuchu\}@mit.edu}}%
\thanks{$^{2} $Dawei Sun is with University of Illinois at Urbana-Champaign, Champaign,
IL, 61820 USA.
        {\tt\footnotesize daweis2@illinois.edu}}%
\thanks{Digital Object Identifier (DOI): see top of this page.}
}

\markboth{IEEE Robotics and Automation Letters. Preprint Version. Accepted January, 2022}
{Qin \MakeLowercase{\textit{et al.}}: \textsc{Sablas}: Learning Safe Control for Black-box Dynamical Systems}

\maketitle

\begin{abstract}
Control certificates based on barrier functions have been a powerful tool to generate probably safe control policies for dynamical systems. However, existing methods based on barrier certificates are normally for white-box systems with differentiable dynamics, which makes them inapplicable to many practical applications where the system is a black-box and cannot be accurately modeled. On the other side, model-free reinforcement learning (RL) methods for black-box systems suffer from lack of safety guarantees and low sampling efficiency. In this paper, we propose a novel method that can learn safe control policies and barrier certificates for black-box dynamical systems, without requiring for an accurate system model. Our method re-designs the loss function to back-propagate gradient to the control policy even when the black-box dynamical system is non-differentiable, and we show that the safety certificates hold on the black-box system. Empirical results in simulation show that our method can significantly improve the performance of the learned policies by achieving nearly 100\% safety and goal reaching rates using much fewer training samples, compared to state-of-the-art black-box safe control methods. Our learned agents can also generalize to unseen scenarios while keeping the original performance. The source code can be found at \url{https://github.com/Zengyi-Qin/bcbf}.

\end{abstract}


\section{Introduction}

\IEEEPARstart{G}{uaranteeing} safety is an open challenge in designing control policies for many autonomous robotic systems, ranging from consumer electronics to self-driving cars and aircrafts. In recent years, the development of machine learning (ML) has created unprecedented opportunities to control modern autonomous systems with growing complexity. However, ML also poses great challenges for developing high-assurance autonomous systems that are provably dependable. While many learning-based approaches~\cite{qin2020keto, qin2021density, schulman2017proximal, lillicrap2015continuous} have been proposed to train controllers to accomplish complex tasks with improved empirical performance, the lack of safety certificates for the learning-enabled components has been a fundamental hurdle that blocks the massive deployment of the learned solutions.

\begin{figure}
    \centering
    \includegraphics[width=0.8\linewidth]{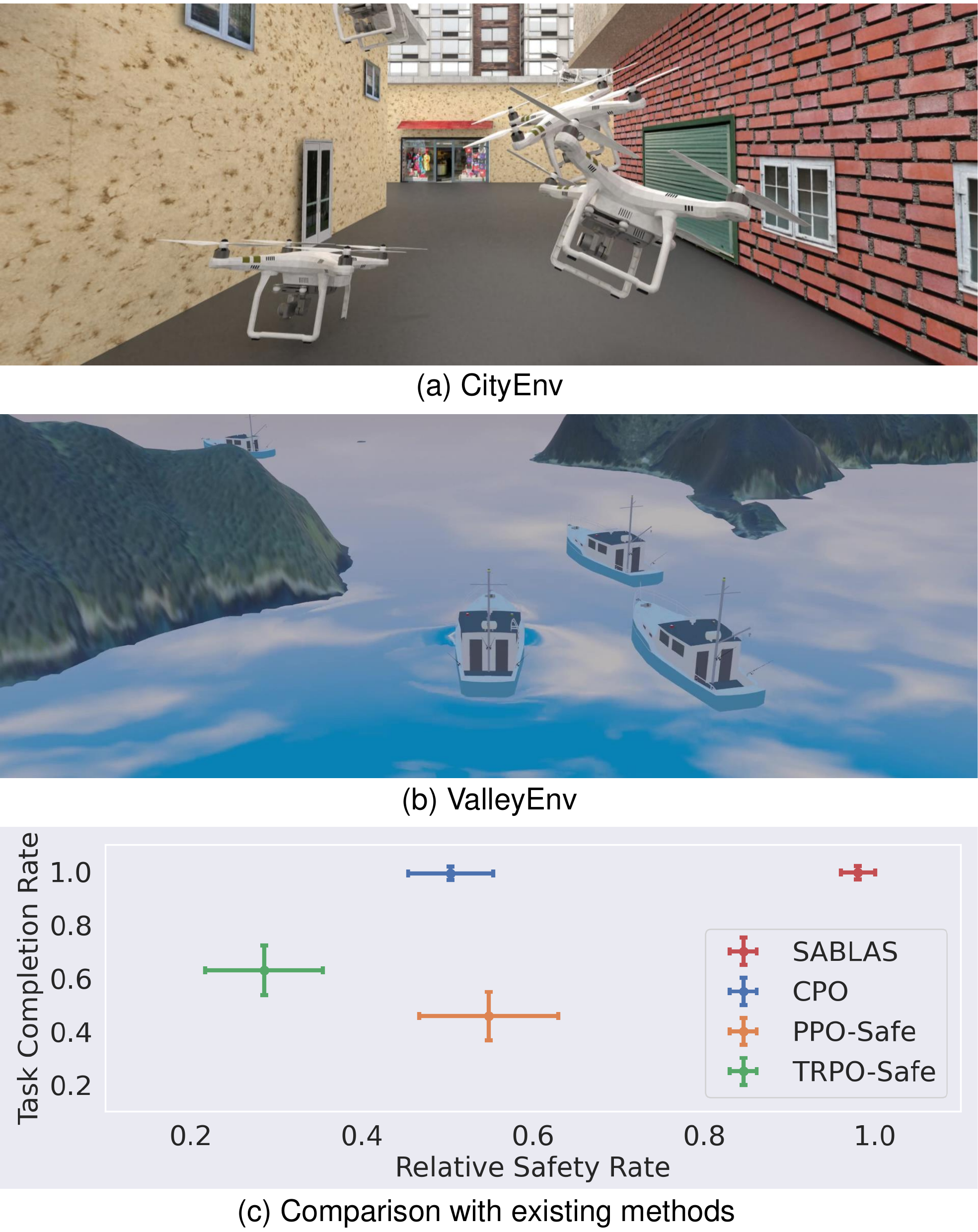}
    \caption{(a) and (b) are two task environments we consider. In both CityEnv and ValleyEnv, the controller should control the ego vehicle to avoid collision with NPC vehicles. There are 1024 NPC drones in the city and 32 NPC ships in the valley. Each vehicle is assigned a random initial and goal location. (c) shows the average result for the two tasks, where the errorbars represent two times the standard deviation. SABLAS exhibits significant improvement in relative safety rate and task completion rate.}
    \label{fig:env_and_result}
\end{figure}

For decades, mathematical control certificates has been used as proofs that the desired properties of the system are satisfied in closed-loop with certain control policies. For example, Control Lyapunov Function~\cite{freeman2008robust, isidori1995nonlinear} and Control Contraction Metrics~\cite{lohmiller1998contraction,manchester2017control,sun2020learning,tsukamoto2020neural} ensure the existence of a controller under which the system converges to an equilibrium or a desired trajectory. Control Barrier Function~\cite{ames2014control,ames2019control,borrmann2015control,chen2020guaranteed,cheng2019end,cheng2020safe, qin2021learning} ensures the existence of a controller that keeps the system inside a safe invariant set. 
Classical control synthesis methods based on Sum-of-Squares~\cite{ames2017control,ames2019control,papachristodoulou2005tutorial,topcu2008local} and linear programs~\cite{kapinski2014simulation,ravanbakhsh2019learning} usually only work for systems with low-dimensional state space and simple dynamics. This is because they choose polynomials as the candidate certificate functions,  where low-order polynomials may not be a valid certificate for complex dynamical systems and high-order polynomials will increase the number of decision variables exponentially in those optimization problems. 
Recent data-driven methods~\cite{qin2021learning,chang2019neural,sun2020learning,dai2021lyapunov} have shown significant progress in overcoming the limitations of the traditional synthesis methods. They can jointly learn the control certificates and policies as neural networks (NN). However, data-driven methods that can generate control certificates still require an explicit (or white-box) differentiable model of the system dynamics, as the derivatives of the dynamics are required in the learning process. 
Finally, many works study the use of control certificates such as CBF to provide safety guarantees during and after training the policies~\cite{cheng2020safe,2020Guaranteeing}, but they more or less requires an accurate model of the system dynamics or a reasonable modeling of the system uncertainties to build the certificates.


Many dynamical systems in the real-world are black-box and lack accurate models. The most popular model-free approach to handle such black-box systems is safe reinforcement learning (RL).
Safe RL methods enforce safety and performance by maximizing the expectation of the cumulative reward and constraining the expectation of the cost to be less or equal to a given threshold. The biggest disadvantage of safe RL methods is the lack of systematic or theoretically grounded way of designing cost function and reward functions, which heavily rely on empirical trials and errors. The lack of explainable safety guarantees and low sampling efficiency also make safe RL methods difficult to exhibit satisfactory performance. 



Instead of stressing about the trade-off between the strong guarantees from control certificates and the practicability of model-free methods, in this work, we propose SABLAS to achieve both. SABLAS is a general-purpose approach to learning \textbf{sa}fe control for \textbf{bla}ck-box dynamical \textbf{s}ystems. SABLAS  enjoys the guarantees provided by the safety certificate from CBF theory without requiring for an accurate model for the dynamics. Instead, SABLAS only needs a nominal dynamics function that can be obtained through regressions over simulation data. There is no need to model the error between the nominal dynamics and the real dynamics since SABLAS essentially re-design the loss function in a novel way to back-propagate gradient to the controller even when the black-box dynamical system is non-differentiable. 
The resulting CBF (and the corresponding safety certificate) holds directly on the original black-box system if the training process converges.
The proposed algorithm is easy-to-implement and follows almost the same procedure of learning CBF for white-box systems with minimal modification.
SABLAS fundamentally solves the problem that control certificates cannot be learned directly on black-box systems, and opens up the next chapter use of CBF theory on synthesizing safe controllers for black-box dynamical systems.


Experimental results demonstrates the superior advantages of SABLAS over leading learning-based safe control methods for black-box systems including CPO~\cite{achiam2017constrained}, PPO-Safe~\cite{schulman2017proximal} and TRPO-Safe~\cite{schulman2015trust, tessler2018reward}. We evaluate SABLAS on two challenging tasks in simulation: drone control in a city and ship control in a valley (as shown in Fig.~\ref{fig:env_and_result}). The dynamics of the drone and ship are assumed unknown. In both tasks, the controlled agent should avoid collision with uncontrolled agents and other obstacles, and reach its goal before the testing episode ends. We also examine the generalization capability of SABLAS on testing scenarios that are not seen in training. Fig.~\ref{fig:env_and_result} shows that SABLAS can reach a near 1.0 relative safety rate and task completion rate while using only $1/10$ of the training data compared to existing safe RL methods, demonstrating a significant improvement. 
We also study the effect of model error (between the nominal model and actual dynamics) on the performance of the learned policy. It is shown that SABLAS is tolerant to large model errors while keeping a high safety rate.
A detailed description of the results is presented in the experiment section. Video results can be found at supplementary materials. 

To summarize the strength of SABLAS:
\begin{inparaenum}
\item SABLAS can jointly find a safety certificate (i.e. CBF) and the corresponding control policy on black-box dynamics;
\item Unlike RL-based methods that need tedious trial-and-error on designing the rewards, SABLAS provides a systematic way of learning certified control policy, without parameters (other than the standard hyper-parameters in NN training) that need fine-tuning. 
\item Empirical results that SABLAS can achieve a nearly perfect performance in terms of guaranteeing safety and goal-reaching, using much less samples than state-of-the-art safe RL methods.
\end{inparaenum}




\section{Related Work}
There is a rich literature on controlling black-box systems and safe RL. Due to the space limit, we only discuss a few directly related and commonly used techniques on black-box system control. We will also skip the literature review for the large body of works on model-based safe control and trajectory planning as the research problems we are solving are very different.
\subsection{Controller Synthesis for Black-box Systems}
Proportional–integral–derivative (PID) controller is widely used in controlling black-box systems. The advantage of PID controller is that it does not rely on a model of the system and only requires the measurement of the state. A drawback of PID controller is that it does not guarantee safety or stability, and the system may overshoot or oscillate about the control setpoint. If the underlying black-box system is linear time-invariant, existing work~\cite{pmlr-v134-chen21c} has presented a polynomial-time control algorithm without relying on any knowledge of the environment. For non-linear black-box systems, the dynamics model can be approximated using system identification and controlled using model-based approaches~\cite{fliess2006complex}, and PID can be used to handle the error part. The concept of practical relative degree~\cite{levant2012practical} is also proposed to enhance the control performance on systems with heavy uncertainties. Recent advance in reinforcement learning~\cite{lillicrap2015continuous, schulman2017proximal, schulman2015trust, grathwohl2017backpropagation} also gives us insight into treating the system as a pure black-box and estimating the gradient for the black-box functions in order to optimize the control variables. However, in safety-critical systems, these black-box control methods still lack formal safety guarantee or certificate.

Simulation-guided controller synthesis methods can also generate control certificates for black-box systems, and sometimes those certificates can indicate how policies should be constructed~\cite{kapinski2014simulation,topcu2008local,ravanbakhsh2019learning}. However, most of these techniques use polynomial templates for the certificates, which limits their use on high-dimensional and complex systems. Another line of work studies the use of ~\cite{djeumou2021fly,djeumou2021fly2} data-driven reachability analysis, jointly with receding-horizon control to constructed optimal control policies. These methods rely on side information about the black-box systems (e.g. lipschitz constants of the dynamics, monotonicity of the states, decoupling in the states' dynamics) to do the reachability, which is not needed in our method.

\subsection{Safe Reinforcement Learning}
Safe RL~\cite{achiam2017constrained, tessler2018reward, qin2021density, yang2020projection, sun2021fisar} extends RL by adding constraints on the expectation of certain cost functions, which encode safety requirements or resource limits. CPO~\cite{achiam2017constrained} derived a policy improvement step that increases the reward while satisfying the safety constraint. DCRL~\cite{qin2021density} imposes constraint on state density functions rather than cost value functions, and shows that density constraint has better expressiveness over cost value function-based constraints. RCPO~\cite{tessler2018reward} weights the cost using Lagrangian multipliers and add it to the reward. FISAR~\cite{sun2021fisar} uses a meta-optimizer to achieve forward invariance in the safe set. A disadvantage of safe RL is that it does not provide safety guarantee, or their safety guarantee cannot be realized in practice. The problem of sampling efficiency and sparsity of cost also increase the difficulty to synthesize safe controller through RL.

\subsection{Safety Certificate and Control Barrier Functions}
Mathematical certificates can serve as proofs that the desired property of the system is satisfied under the corresponding planning~\cite{tordesillas2019faster, tordesillas2021mader} and control components. Such certificate is able to guide the controller synthesis for dynamical systems in order to ensure safety. For example, Control Lyapunov Function~\cite{freeman2008robust, isidori1995nonlinear, dai2021lyapunov} ensures the existence of a controller so that the system converges to desired behavior. Control Barrier Function~\cite{ames2014control,ames2019control,borrmann2015control,chen2020guaranteed,cheng2019end,cheng2020safe, qin2021learning, choi2021robust} ensures the existence of a controller that keep the system inside a safe invariant set. However, the existing controller synthesis with safety certificate heavily relies on a white-box model of the system dynamics. For black-box systems or system with large model uncertainty, these methods are not applicable. While recent work~\cite{castaneda2021pointwise} proposes to learn the model uncertainty effect on the CBF conditions, it still assumes that a handcrafted CBF is given, which is not always available for complex non-linear dynamical systems. Our approach represents a substantial improvement over the existing CBF-based safe controller synthesis strategy. The proposed SABLAS framework simultaneously enjoys the safety certificate from the CBF theory and the effectiveness on black-box dynamical systems.
\section{Preliminaries}

\subsection{Safety of Black-box Dynamical Systems}

\begin{definition}[Black-box Dynamical System]\label{def:dynamical_system}
A black-box dynamical system is represented by tuple $\langle \mathcal{S}, \mathcal{U}, f \rangle$, where $\mathcal{S}\in \mathbb{R}^n$ is the state space and $\mathcal{U}\in \mathbb{R}^m$ is the control input space. $f: \mathcal{S} \times \mathcal{U} \mapsto \mathcal{S}$ is the system dynamics $\dot{s} = f(s, u)$, which is unknown due to the black-box assumption.
\end{definition}

Let $\mathcal{S}_0, \mathcal{S}_g, \mathcal{S}_s$ and $\mathcal{S}_d$ denote the set of initial states, goal states, safe states and dangerous states respectively. The problem we aim to solve is formalized in Definition~\ref{def:safe_control}.

\begin{definition}[Safe Control of Black-box Systems]\label{def:safe_control}
Given a black-box dynamical system modeled as in Definition~\ref{def:dynamical_system}, the safe control problem aims to find a controller $\pi: \mathcal{S}\mapsto \mathcal{U}$ such that under control input $u = \pi(s)$ and the unknown dynamics $\dot{s} = f(s, u)$, the following is satisfied:
\begin{equation}
    \begin{aligned}
    &\exists~t > 0,~s(t)\in \mathcal{S}_g \\
    &\text{s.t.}~\forall~t>0,~s(t) \not\in \mathcal{S}_d \text{~and~} s(0)\in \mathcal{S}_0
    \end{aligned}
\end{equation}
\end{definition}

The above definition requires that starting from the initial set $\mathcal{S}_0$, the system should never enter the dangerous set $\mathcal{S}_d$ under controller $\pi$.

\subsection{Control Barrier Function as Safety Certificate}
A common approach for guaranteeing safety of dynamical systems is via control barrier functions (CBF)~\cite{ames2014control}, which ensures that the state always stay in the safe invariant set. A control barrier function $h: \mathcal{S} \mapsto \mathbb{R}$ satisfies:
\begin{equation}\label{eqn:cbf_conditions}
    \begin{aligned}
    & h(s) \geq 0, 
    \quad \quad \quad \quad \quad \quad \quad \quad \quad \quad \quad~~ 
    \forall s \in \mathcal{S}_0 \\
    & h(s) < 0, 
    \quad \quad \quad \quad \quad \quad \quad \quad \quad \quad \quad~~
    \forall s \in \mathcal{S}_d \\
    & \dot{h} + \alpha(h) = \frac{\partial h}{\partial s} f(s, u) + \alpha(h) \geq 0, \quad\forall s \in \mathcal{S}_p
    \end{aligned}
\end{equation}
where $\mathcal{S}_p = \{s\mid h(s)\geq 0\}$, $u=\pi(s)$ and $\alpha(\cdot)$ is a class-$\mathcal{K}$ function that is strictly increasing and $\alpha(0)=0$. It is proven~\cite{ames2014control} that if there exists a CBF $h$ for a given controller $\pi$, the system controlled by $\pi$ starting from $s(0)\in \mathcal{S}_0$ will never enter the dangerous set $\mathcal{S}_d$. Besides the formal safety proof in \cite{ames2014control}, there is an informal but straightforward way to understand the safety guarantee. Whenever $h(s)$ decreases to $0$, we have $\dot{h} + \alpha(h) = \dot{h} \geq 0$, which means $h(s)$ no longer decreases and $h(s) < 0$ will not occur. Thus $s$ will not enter the dangerous set $\mathcal{S}_d$ where $h(s) < 0$.

\subsection{Co-learning Controller and CBF for White-box Systems}
For white-box dynamical systems where $f(s, u)$ is known, we can jointly synthesize the controller $\pi$ and it safety certificate $h$ that satisfy CBF conditions~\eqref{eqn:cbf_conditions} via a learning-based approach. We model $\pi$ and $h$ as neural networks with parameters $\Theta$ and $\Omega$. Given a dataset $\mathcal{D}$ of the state samples in $\mathcal{S}$, the CBF conditions~\eqref{eqn:cbf_conditions} can be translated into empirical loss functions:
\begin{equation} \label{eqn:loss_white_box}
    \begin{aligned}
    & \mathcal{L}_0(\Omega) = \frac{1}{|\mathcal{S}_0\cap \mathcal{D}|}\sum_{s\in \mathcal{S}_0\cap \mathcal{D}} \max(0, -h(s; \Omega)) \\
    & \mathcal{L}_d(\Omega) = \frac{1}{|\mathcal{S}_d\cap \mathcal{D}|}\sum_{s\in \mathcal{S}_d\cap \mathcal{D}} \max(0, h(s; \Omega)) \\
    & \mathcal{L}_p(\Omega, \Theta) = \frac{1}{|\mathcal{S}_p\cap \mathcal{D}|} \sum_{s \in \mathcal{S}_p\cap \mathcal{D}} \\ 
    & \quad \quad \quad \quad \max\left(0, -\frac{\partial h}{\partial s} f(s, \pi(s; \Theta)) - \alpha(h(s, \Omega))\right). \\
    \end{aligned}
\end{equation}
Each of the loss functions in \eqref{eqn:loss_white_box} corresponds to a CBF condition in \eqref{eqn:cbf_conditions}. In addition to safety, we also consider goal-reaching by penalize the difference between the safe controller $\pi$ and the goal-reaching nominal controller $\pi_{nom}$ in \eqref{eqn:loss_goal_reaching}. The synthesis of $\pi_{nom}$ is well-studied and is not the contribution of our work.
\begin{equation}\label{eqn:loss_goal_reaching}
    \begin{aligned}
    \mathcal{L}_{g}(\Theta) = \frac{1}{|\mathcal{S}_p\cap \mathcal{D}|} \sum_{s \in \mathcal{S}_p\cap \mathcal{D}} ||\pi(s; \Theta) - \pi_{nom}(s)||_2^2.
    \end{aligned}
\end{equation}
The total loss function is $\mathcal{L}_0(\Omega) + \mathcal{L}_d(\Omega) + \mathcal{L}_p(\Omega, \Theta) + \lambda \mathcal{L}_{g}(\Theta)$, where $\lambda$ is a constant that balances the goal-reaching and safety objective. The total loss function is minimized via stochastic gradient descent to find the parameters $\Omega$ and $\Theta$. The dataset $\mathcal{D}$ is not fixed during training and will be periodically updated with new samples by running the current controller. When the loss $\mathcal{L}_0+\mathcal{L}_d+\mathcal{L}_p$ converges to $0$, the resulting controller $\pi(s; \Theta)$ and CBF $h(s; \Omega)$ will satisfy \eqref{eqn:cbf_conditions} on unseen testing samples with a generalization bound as proven in~\cite{qin2021learning}. Therefore, a safe controller and the corresponding CBF is found for the white-box system.

However, for black-box dynamical systems where $f(s, u)$ is unknown, the co-learning method for safe controller and CBF described above is no longer applicable. In Section~\ref{sec:approach}, we propose an important and easy-to-implement modification to~\eqref{eqn:loss_white_box} such that we can leverage similar co-learning framework to jointly synthesize the safe controller and its corresponding CBF as safety certificate.
\section{Learning CBF on black-box systems} \label{sec:approach}

In this section, we first elaborate on why it is difficult to learn the safe controller and its CBF for black-box dynamical systems. Then we will propose an important and easy-to-implement re-formulation of the optimization objective, which makes learning safe controller in black-box systems as easy as in white-box systems.

\subsection{Challenges in Black-box Dynamical Systems}\label{sec:challenges}
Among the three loss functions $\mathcal{L}_0, \mathcal{L}_d~\text{and}~\mathcal{L}_p$ in \eqref{eqn:loss_white_box}, $\mathcal{L}_p$ is the only one that can propagate gradient to the controller parameter $\Theta$. The main challenge of training safe controller with its CBF for black-box dynamical systems is that the gradient can no longer be back-propagated to $\Theta$ when $f$ is unknown. Therefore, a safe controller cannot be trained by minimizing the loss functions in \eqref{eqn:loss_white_box}.

Given state samples $s(t)$ and $s(t+\Delta t)$ from the black-box system where $\Delta t$ is a sufficiently small time interval, we can approximate $\dot{h}$ and compute $\mathcal{L}_p$ as:
\begin{equation} \label{eqn:loss_only_h}
    \begin{aligned}
    &\dot{h}_{1}(s) = \frac{h(s(t+\Delta t)) - h(s(t))}{\Delta t} \\
    &\mathcal{L}_{p1} = \frac{1}{|\mathcal{S}_p\cap \mathcal{D}|} \sum_{s \in \mathcal{S}_p\cap \mathcal{D}} \max\left(0, -\dot{h}_{1}(s) - \alpha(h(s))\right).
    \end{aligned}
\end{equation}
$\mathcal{L}_{p1}$ does give the value of $\mathcal{L}_p$, but its backward gradient flow to the controller is cut off by the black-box system that is non-differentiable. \eqref{eqn:loss_only_h} can only be used to train CBF $h$ but not the safe controller $\pi$. Even worse, the $h$ obtained by minimizing \eqref{eqn:loss_only_h} does not guarantee that a corresponding safe controller $\pi$ exists. If we have an differential expression of dynamics $f$ and replace $h(s(t+\Delta t))$ with $h(s(t) + f(s(t), \pi(s(t)))\Delta t)$, the gradient flow can successfully reach $\pi(s)$ and update the controller parameter. However, this is not immediately possible because $f$ is unknown by the black-box assumption.

A possible way to back-propagate gradient to $\pi$ is to use a differentiable nominal model $f_{nom}$. There are many methods to obtain $f_{nom}$, such as fitting a neural network using sampled data from the real black-box system. We do not require $f_{nom}$ to perfectly match the real dynamics $f$, because there will always exist an error between them. With $f_{nom}$, we can approximate $\mathcal{L}_p$ as:
\begin{equation} \label{eqn:loss_fnom}
    \begin{aligned}
    &\dot{h}_2(s) = \frac{h(s + f_{nom}(s, \pi(s))\Delta t) - h(s)}{\Delta t}\\
    &\mathcal{L}_{p2} = \frac{1}{|\mathcal{S}_p\cap \mathcal{D}|} \sum_{s \in \mathcal{S}_p\cap \mathcal{D}} \max\left(0, -\dot{h}_2(s) - \alpha(h(s))\right),
    \end{aligned}
\end{equation}
which is differentiable w.r.t. $\pi(s)$ because both $h$ and $f_{nom}$ are differentiable. The gradient of $\mathcal{L}_{p2}$ can be back-propagated to the controller to update its parameters. However, whatever way we get $f_{nom}$, there still exists an error between the real dynamics $f$ and $f_{nom}$, which means $\dot{h}_2$ is not a good approximation of $\dot{h}$ and $\mathcal{L}_{p2}$ is not the true value of $\mathcal{L}_p$. Using $\mathcal{L}_{p2}$, it is not guaranteed that the third CBF condition in \eqref{eqn:cbf_conditions} will be satisfied.

\subsection{Learning Safe Control for Black-box Dynamics}
\begin{figure}
    \centering
    \includegraphics[width=0.8\linewidth]{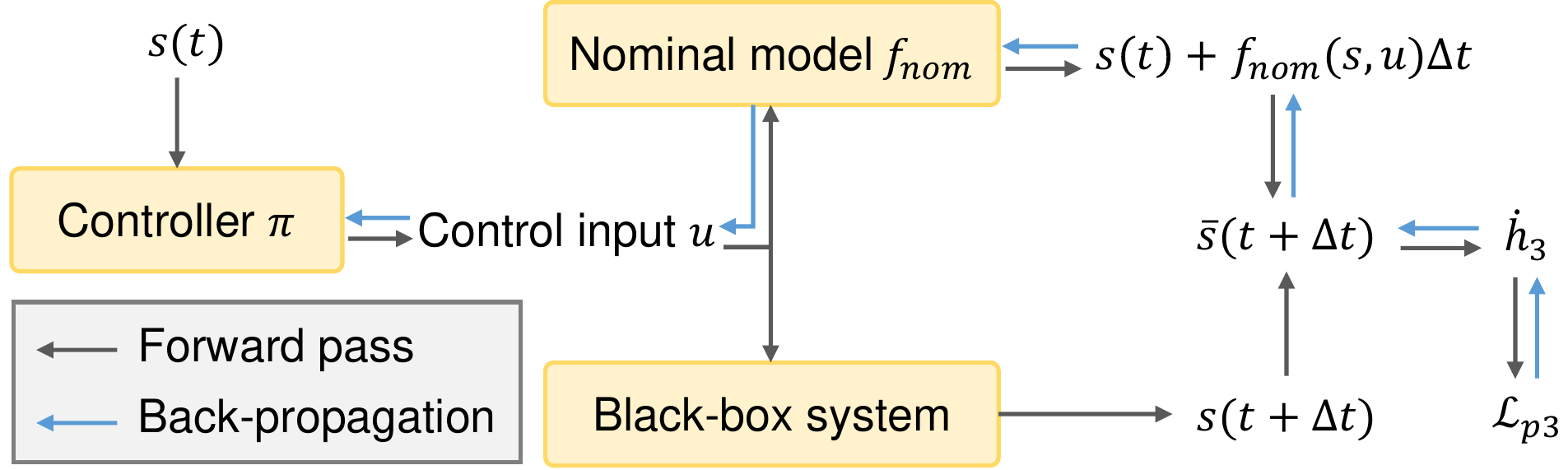}
    \caption{Computational graph of $\mathcal{L}_{p3}$. The blue lines show how the gradient from $\mathcal{L}_{p3}$ is back-propagated to the controller $\pi$ despite that the black-box system itself is non-differentiable.}
    \label{fig:computational_graph}
\end{figure}
We present a novel re-formulation of $\mathcal{L}_p$ that makes learning safe controller with CBF for black-box dynamical systems as easy as in white-box systems. The proposed formulation possesses two features: it enables the gradient to back-propagate to the controller in training, and offers an error-free approximation of $\dot{h}$.

Given state samples $s(t)$ and $s(t+\Delta t)$ from the trajectories of the real black-box dynamical system, where $\Delta t$ is a sufficiently small time interval, we define $s_{nom}(t+\Delta t)$ as :
\begin{equation}
    \begin{aligned}
    s_{nom}(t+\Delta t) = s(t) + f_{nom}(s(t), \pi(s(t)))\Delta t, \nonumber
    \end{aligned}
\end{equation}
then construct $\bar{s}(t+\Delta t)$ as:
\begin{equation}
    \begin{aligned}
    \bar{s}(t+\Delta t) =  s_{nom}(t+\Delta t) + g(s(t+\Delta t) - s_{nom}(t+\Delta t)), \nonumber
    \end{aligned}
\end{equation}
where $g(s) = s$ is an identity function but \textit{without gradient}. We need to pretend that $g(s)$ is a constant and in back-propagation, the gradient on $g(s)$ cannot propagate to its argument $s$. In PyTorch~\cite{paszke2019pytorch}, there is an off-the-shelf implementation of $g(s)$ as $g(s) = s.\rm{detach()}$, which cuts off the gradient from $g$ to $s$ in back-propagation. Then we approximate $\mathcal{L}_p$ using:
\begin{equation}\label{eqn:loss_Lp3}
    \begin{aligned}
    &\dot{h}_{3}(s) = \frac{h(\bar{s}(t+\Delta t)) - h(s(t))}{\Delta t} \\
    &\mathcal{L}_{p3} = \frac{1}{|\mathcal{S}_p\cap \mathcal{D}|} \sum_{s \in \mathcal{S}_p\cap \mathcal{D}} \max\left(0, -\dot{h}_{3}(s) - \alpha(h(s))\right).
    \end{aligned}
\end{equation}
\begin{theorem}\label{theo:main_theorem}
$\nabla_{\Theta}\mathcal{L}_{p3}$ exists and $\lim_{\Delta t\rightarrow 0} \mathcal{L}_{p3} = \mathcal{L}_p$. Namely, $\mathcal{L}_{p3}$ is differentiable w.r.t. the controller parameter $\Theta$, and $\mathcal{L}_{p3}$ is an error-free approximation of $\mathcal{L}_p$ as $\Delta t \rightarrow 0$.
\end{theorem}
\begin{proof}
Since $f_{nom}$ is differentiable, $s_{nom}$ and $\bar{s}(t+\Delta t)$ are differentiable w.r.t. $\pi$, $\dot{h}_3$ and $\mathcal{L}_{p3}$ are also differentiable w.r.t. $\pi$ and its parameter $\Theta$. Thus, $\nabla_{\Theta}\mathcal{L}_{p3}$ exists. Furthermore, since $\bar{s}(t+\Delta t)=s(t+\Delta t)$, $\dot{h}_3$ is an error-free approximation of the real $\dot{h}$ when $\Delta t\rightarrow 0$. Thus $\mathcal{L}_{p3}$ is also an error-free approximation of $\mathcal{L}_p$ when $\Delta t\rightarrow 0$.
\end{proof}

Note that Theorem~\ref{theo:main_theorem} reveals the reason why the proposed SABLAS method can jointly learn the CBF and the safe controller for black-box dynamical system. \textbf{First}, since $\nabla_{\Theta}\mathcal{L}_{p3}$ exists, the gradient from $\mathcal{L}_{p3}$ can be back-propagated to the controller parameters to learn a safe controller. On the contrary, $\dot{h}_1$ is not differentiable w.r.t. $\pi$ so $\mathcal{L}_{p1}$ cannot be used to train the controller. \textbf{Second}, since $\mathcal{L}_{p3}$ is a good approximation of $\mathcal{L}_p$, minimizing $\mathcal{L}_{p3}$ contributes to the minimization of $\mathcal{L}_{p}$ and the satisfaction of the third CBF condition in \eqref{eqn:cbf_conditions}. On the contrary, $\dot{h}_2$ is an inaccurate approximation of $\dot{h}$ as we elaborated in Section~\ref{sec:challenges}. The construction of $\mathcal{L}_{p3}$ incorporates the advantages of $\mathcal{L}_{p1}$ and $\mathcal{L}_{p2}$, and avoids the disadvantages of $\mathcal{L}_{p1}$ and $\mathcal{L}_{p2}$ at the same time. The computational graph of $\mathcal{L}_{p3}$ is illustrated in Fig.~\ref{fig:computational_graph}, which shows the forward pass and the backward gradient propagation from $\mathcal{L}_{p3}$ to controller $\pi$. 

\textbf{Remark 1.} One may argue that the gradient received by $\pi$ via minimizing $\mathcal{L}_{p3}$ is not exactly the gradient it should receive if we had a perfect differentiable model of the black-box system. Despite this, minimizing $\mathcal{L}_{p3}$ directly contributes to the satisfaction of the third CBF condition in \eqref{eqn:cbf_conditions}. A safe controller and its CBF can be found to keep the system within safe set.

\textbf{Remark 2.} Although the current formulation of $\mathcal{L}_{p3}$ leads to promising performance in simulation as we will show in experiments, $\mathcal{L}_{p3}$ requires further consideration in hardware experiments. Directly using the future state $s(t+\Delta t)$ to calculate the time derivative of $h$ or $s$ is not always desirable because noise will possibly dominate the numerical differentiation. When the noise dominates the time derivative of $s$ or $h$, the training will have convergence issues. But a moderate noise is actually beneficial to training, because our optimization objective makes the CBF conditions hold even under noise disturbance, which increases the robustness of the trained CBF and controller. On physical robots where noise dominates the numerical differentiation, one can incoperate filtering techniques to mitigate the noise.

Combining with the loss functions $\mathcal{L}_0, \mathcal{L}_d$ and $\mathcal{L}_g$ in \eqref{eqn:loss_white_box} and \eqref{eqn:loss_goal_reaching}, the total loss function can be formulated as:
\begin{equation}
    \begin{aligned}
    \mathcal{L}(\Omega, \Theta) = \mathcal{L}_0(\Omega) + \mathcal{L}_d(\Omega) + \mathcal{L}_{p3}(\Omega, \Theta) + \lambda \mathcal{L}_{g}(\Theta),
    \end{aligned}
\end{equation}
where $\lambda$ is a constant balancing the safety and goal-reaching objective. $\mathcal{L}$ is minimized via stochastic gradient descent. Algorithm~\ref{alg:learn_cbf} summarizes the learning process of the safe controller and the corresponding safety certificate (CBF) for black-box dynamical systems. The \textsc{Run} function runs the black-box system using the controller $\pi$ under initial condition $s(0)$, and returns the trajectory data. The \textsc{Update} function updates parameters $\Omega, \Theta$ by minimizing $\mathcal{L}(\Omega, \Theta)$ using the state samples in $\mathcal{D}$ via gradient descent.

\begin{algorithm}[t]
\caption{Learning Safe Controller with CBF for Black-box Dynamical Systems}
\begin{algorithmic}[1]
\State \textbf{Input:} The set of initial condition $\mathcal{S}_0$, the set of dangerous state $\mathcal{S}_d$, controller $\pi$ and CBF $h$ with randomly initialized parameters $\Theta$ and $\Omega$, nominal dynamics $f_{nom}$, loss function $\mathcal{L}(\Omega, \Theta)$, the number of episodes $K$ in data collection, the number of training iterations $N$.
\For{$i=1, 2, \cdots, N$}
  \State Initialize state samples dataset $\mathcal{D}\leftarrow \emptyset$
  \For{$j=1, 2, \cdots, K$}
  \State $s(0)\leftarrow$ \textsc{Sample}($\mathcal{S}_0$)
  \State $\mathcal{D}_j \leftarrow$ \textsc{Run}$(\pi(\cdot; \Theta), s(0))$, $~\mathcal{D} \leftarrow \mathcal{D} \cup \mathcal{D}_j$
  \EndFor
  \State $\Omega, \Theta \leftarrow$ \textsc{Update}$(\Omega, \Theta, \mathcal{D}, \mathcal{L})$
\EndFor \\
\Return{CBF parameter $\Omega$ and controller parameter $\Theta$}
\end{algorithmic}
\label{alg:learn_cbf}
\end{algorithm}
\section{Experiment}


\begin{figure*}[!ht]
    \centering
    \includegraphics[width=0.8\linewidth]{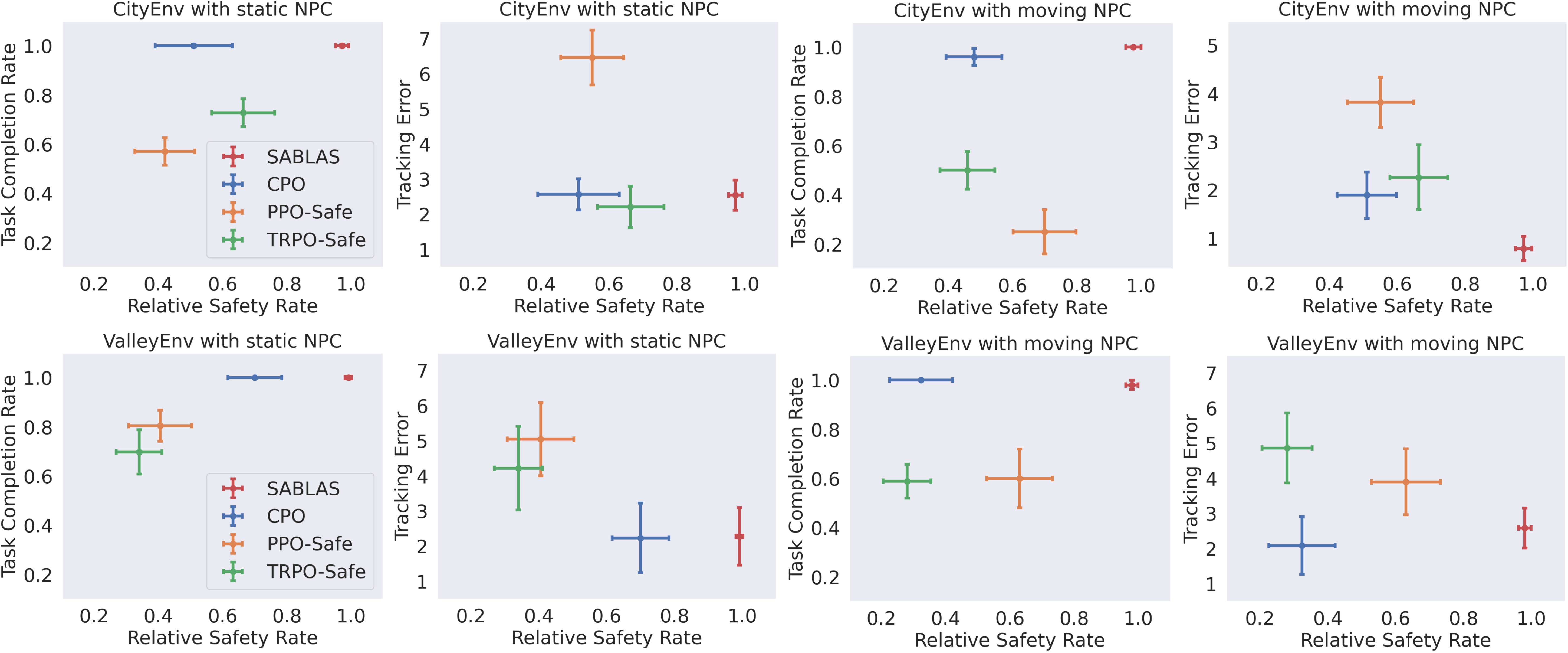}
    \caption{Quantitative results of relative safety rate, task completion rate and tracking error. The center points are mean  and error bars represent standard deviations. Results are collected over 50 random runs on each method. The first row is for drone control in CityEnv and the second row is for ship control in ValleyEnv. Our method reaches a high task completion rate and relative rate at the same time, while keeping a low tracking error.}
    \label{fig:safety_rate_result}
\end{figure*}

The primary objective of our experiment is to examine the effectiveness of the proposed method in terms of safety and goal-reaching when controlling black-box dynamical systems. We will conduct comprehensive experiments on two simulation environments illustrated in Fig.\ref{fig:env_and_result}~(a) and (b), and compare with state-of-the-art learning-based control methods for black-box systems.

\subsection{Task Environment Description}
\paragraph{Drone control in a city (CityEnv)} In our first case study, we consider the package delivery task in a city using drones, as is illustrated in Fig.~\ref{fig:env_and_result}~(a). There is one controlled drone and 1024 non-player character (NPC) drones that are not controlled by our controller. In each simulation episode, each drone is assigned a sequence of randomly selected goals to visit. The aim of our controller is to make sure the controlled drone reach its goals while avoiding collision with NPC drones at any time. A reference trajectory $s_{ref}(t), t\in[0, T]$ will be given, which sequentially connects the goals and avoid collision with buildings. The reference trajectory can be generated by any off-the-shelf single-agent path planning algorithm. We use FACTEST~\cite{FACTEST} in our implementation, and other options such as RRT~\cite{lavalle2001rapidly} are also suitable. The reference path planner does not need to consider the dynamic obstacles, such as the moving NPCs in our experiment. A nominal controller $\pi_{norm}$ will also be given, which outputs control commands that drive the drone to follow the reference trajectory. However, $\pi_{norm}$ is purely for goal-reaching and does not consider safety. The CityEnv has two modes: with \textbf{static} NPCs and \textbf{moving} NPCs. If the NPCs are static, they will constantly stay at their initial locations. If the NPCs are moving, they will follow pre-planned trajectories to sequentially visit their goals. The drone model is with state space $[x, y, z, v_x, v_y, v_z, \theta_x, \theta_y]$, where $\theta_x$ and $\theta_y$ are row and pitch angles. The control inputs are the angular acceleration of $\theta_x, \theta_y$ and the vertical thrust. The underlying model dynamics is from \cite{qin2021learning} and assumed unknown to the controller and CBF in our experiment.

\paragraph{Ship control in a valley (ValleyEnv)} In our second case study, we consider task of controlling a ship in valley illustrated in Fig.~\ref{fig:env_and_result}~(b). There are one controlled ship and 32 NPC ships. The number of NPCs in ValleyEnv is less than CityEnv because ships are large in size and inertia, and hard to maneuver in dense traffic. Also, different from the 3D CityEnv, ValleyEnv is in 2D, which means the agents have fewer degrees of freedom to avoid collision. The initial location and goal location of each ship is randomly initialized at the beginning of each episode. The aim of our controller is to ensure the controlled ship reach its goal and avoid collision with NPC ships. Similar to CityEnv, a reference trajectory and nominal controller will be provided. There are also two modes in ValleyEnv, including the static and moving NPC mode, as is in CityEnv. The ship model is with state space $[x, y, \theta, u, v, \omega]$, where $\theta$ is the heading angle, $u, v$ are speed in the ship body coordinates, and $\omega$ is the angular velocity of the heading angle. The ship model is from Sec.~4.2 of \cite{FOSSEN20001} and is unknown to the controller and CBF.

\subsection{Evaluation Criteria}
Three evaluation criteria are considered. \textbf{Relative safety rate} measures the improvement of safety comparing to a nominal controller that only targets at goal-reaching but not safety. To formally define the relative safety rate, we first consider the absolute safety rate $\alpha$
as $\alpha = \frac{1}{T}\int_0^T\mathbb{I}(s(t) \not\in \mathcal{S}_d)~dt$, which measures the proportion of time that the system stays outside the dangerous set. Given two control policies $\pi_1$ and $\pi_2$ with absolute safety rate $\alpha_1$ and $\alpha_2$, the relative safety rate of $\pi_1$ w.r.t. $\pi_2$ is defined as $\beta_{12} = \frac{\alpha_1 - \alpha_2}{1-\alpha_2} \in (-\infty, 1].$
If $\beta_{12} = 0$, then control policy $\pi_1$ does not have any improvement over $\pi_2$ in terms of safety. If $\beta_{12} = 1$, then $\pi_1$ completely guarantees safety of the system in $t\in[0, T]$. In our experiment, $\pi_1$ is the controller to be evaluated, and $\pi_2$ is the nominal controller $\pi_{norm}$ that only accounts for goal-reaching without considering safety. \textbf{Task completion rate} is defined as the success rate of reaching the goal state before timeout. \textbf{Tracking error} is the average deviation of the system's state trajectory comparing to a pre-planned reference trajectory $s_{ref}(t), t\in[0, T]$ as $\gamma = \frac{1}{T}\int_0^T||s(t) - s_{ref}(t)||^2_2~dt.$
Note that we do not assume $s_{ref}$ always stay outside the dangerous set.

\subsection{Baseline Approaches}
In terms of safe control for black-box systems, the most recent state-of-the-art approaches are safe reinforcement learning (safe RL) algorithms. We choose three safe RL algorithms for comparison: \textbf{CPO}~\cite{achiam2017constrained} is a general-purpose policy optimization algorithm for black-box systems that maximizes the expected reward while satisfying the safety constraints. \textbf{PPO-Safe} is a combination of PPO~\cite{schulman2017proximal} and RCPO~\cite{tessler2018reward}. It uses PPO to maximize the expected cumulative reward while leveraging the Lagrangian multiplier update rule in RCPO to enforce the safety constraint. \textbf{TRPO-Safe} is a combination of TRPO~\cite{schulman2015trust} and RCPO~\cite{tessler2018reward}. The expected reward is maximized via TRPO and the safety constraints are imposed using the Lagrangian multiplier in RCPO. 

\subsection{Implementation and Training} \label{sec:implementation_and_training}
Both the controller $\pi$ and CBF $h$ are multi-layer perceptrons (MLP) with architecture adopted from Sec.~4.2 of~\cite{qin2021learning}. $\pi$ and $h$ not only take the state of the controlled agent as input, but also the states of 8 nearest NPCs that the controlled agent can observe. In Algorithm~\ref{alg:learn_cbf}, we choose $K=1, N=2000$. The total number of state samples collected during training is $10^6$. In \textsc{Update} of the algorithm, we use the Adam~\cite{kingma2014adam} optimizer with learning rate $10^{-4}$ and batch size 1024. The gradient descent runs for 100 iterations in \textsc{Update}. The nominal model dynamics are fitted from trajectory data in simulation. We used $10^4$ state samples to fit a linear approximation of the drone dynamics, and $10^5$ samples to fit a non-linear 3-layer MLP as the ship dynamics. 

In training the safe RL methods, the reward in every step is the negative distance between the system's current state and the goal state, and the cost is 1 if the system is within the dangerous set $\mathcal{S}_d$ and 0 otherwise. The threshold for expected cost is set to 0, which means we wish the system never enter the dangerous set (never reach a state with a positive cost). During training, the agent runs the system for $10^7$ timesteps in total and performs $2000$ policy updates. In each policy update, 100 iterations of gradient descent are performed. The implementation of the safe RL methods is based on~\cite{ray2019benchmarking}.

All the methods are trained with static NPCs and tested on both static and moving NPCs. We believe this can make the testing more challenging and examine the generalization capability of the tested methods in different scenarios. All the agents are assigned random initial and goal locations in every simulation episode, which prevents the learned controller from overfitting a single configuration.

\begin{figure}
    \centering
    \includegraphics[width=0.8\linewidth]{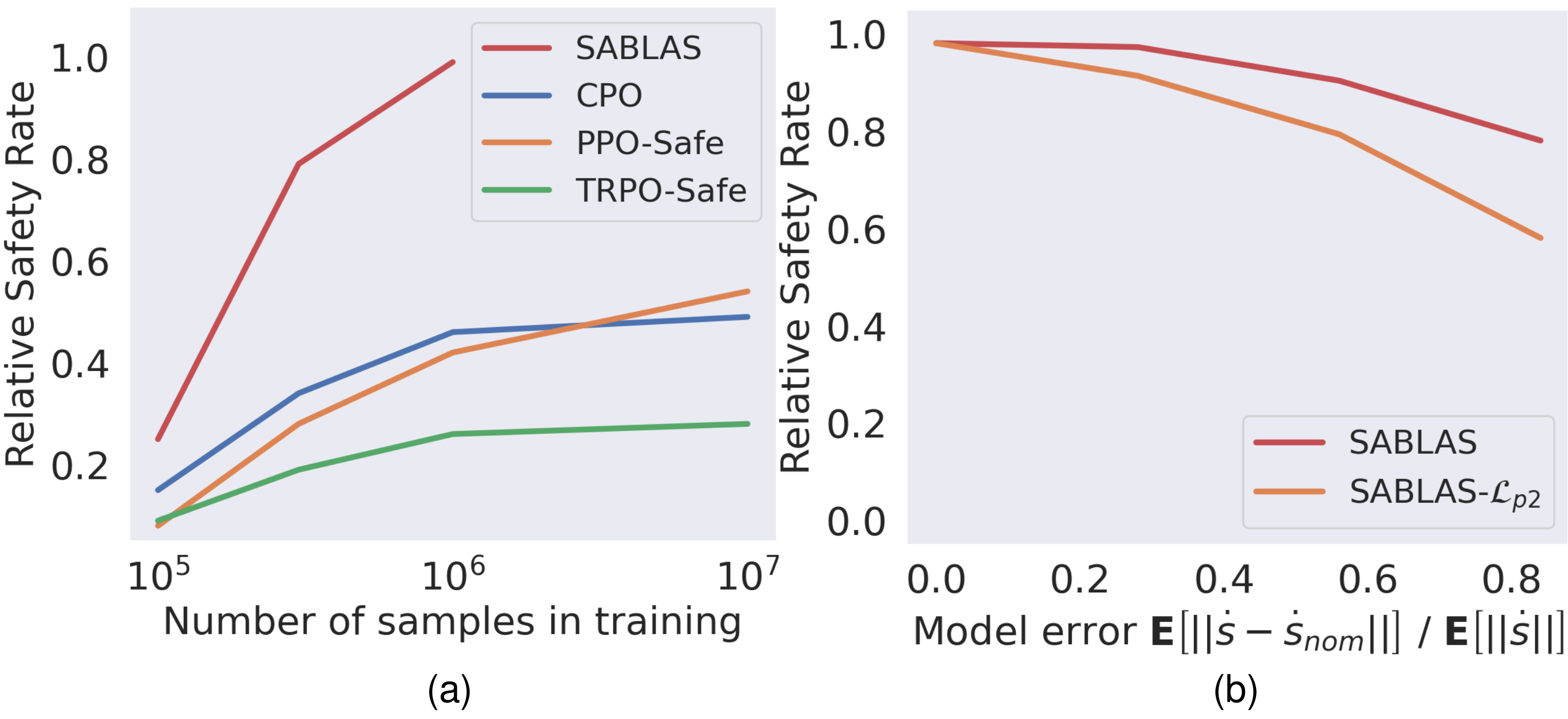}
    \caption{(a) Comparison of sampling efficiency. With only $1/10$ samples, SABLAS achieves a nearly perfect relative safety rate. (b) Influence of model error on safety performance. SABLAS-$\mathcal{L}_{p2}$ uses Eqn.~\ref{eqn:loss_fnom} instead of the proposed Eqn.~\ref{eqn:loss_Lp3} as loss function. SABLAS is tolerant to large error between nominal and real models while keeping a high safety rate.}
    \label{fig:error_num_samples_safety_rate}
\end{figure}

\subsection{Experimental Results}
\paragraph{Safety and goal-reaching performance} Results are shown in Fig.~\ref{fig:safety_rate_result}. Among the compared methods, our method is the only one that can reach a high task completion rate and relative safety rate at the same time. For other methods such as TRPO-Safe, when the controlled drone or ship is about to hit the NPCs, the learned controller tend to brake and decelerate. Thus, the agent is less likely to reach its goal when a simulation episode ends. The task completion rate and safety rate are opposite to each other for CPO, PPO-Safe and TRPO-Safe. On the contrary, the controller obtained by our method can maneuver smoothly among NPCs without severe deceleration. This enables the controlled agent to reach the goal location on time. Our method can also keep a relatively low tracking error, which means the difference between actual trajectories and reference trajectories is small.

\paragraph{Generalization capability to unseen scenarios} As is stated in Sec.~\ref{sec:implementation_and_training}, the NPCs are static in training and can be either static or moving in testing. Fig.~\ref{fig:safety_rate_result} also demonstrates that our method has promising generalization capability across different training and testing scenarios.

\paragraph{Sampling efficiency} In Fig.~\ref{fig:error_num_samples_safety_rate}~(a), we show the safety performance under different sizes of the training set. The results are averaged over the drone and ship control tasks. SABLAS only needs around $1/10$ of the samples required by the compared methods to achieve a nearly perfect relative safety rate. Note that SABLAS requires an extra $10^4$ to $10^5$ samples to fit the nominal dynamics, but this does not change the fact that the total number of samples needed by SABLAS is much fewer than the baselines.

\paragraph{Effect of model error} We investigate the influence of the model error $||\dot{s}-\dot{s}_{nom}||$ between the real dynamics and the nominal dynamics on the safety performance. We change the modeling error of the drone model and test the learning controller on CityEnv with static NPCs. We also perform an ablation study where we use $\mathcal{L}_{p2}$ in Eqn.~\ref{eqn:loss_fnom} instead of $\mathcal{L}_{p3}$ in Eqn.~\ref{eqn:loss_Lp3} as loss function.  The red curve in Fig.~\ref{fig:error_num_samples_safety_rate}~(b) show that SABLAS is tolerant to large model errors while exhibiting a promising safety rate. In our previous experiments, the model error $e = \mathbf{E}[||\dot{s}-\dot{s}_{nom}||] / \mathbf{E}[||\dot{s}||]$ is always less than $0.2$. We did not encounter any difficulty fitting a nominal model with empirical error $e\leq 0.2$.
The orange curve in Fig.~\ref{fig:error_num_samples_safety_rate}~(b) shows that if we use $\mathcal{L}_{p2}$ in Eqn.~\ref{eqn:loss_fnom}, the trained controller will have a worse performance in terms of safety rate. This is because $\mathcal{L}_{p2}$ only uses the nominal dynamics to calculate the loss, without leveraging the real black-box dynamics.

\subsection{Discussion on Limitation}
The main limitation of the proposed approach is that it cannot guarantee the satisfaction of the CBF conditions in \eqref{eqn:cbf_conditions} in the entire state space. Even if we minimize $\mathcal{L}_0, \mathcal{L}_d$ and $\mathcal{L}_{p3}$ to $0$ during training, the CBF conditions may still be occasionally violated during testing. After all, the training samples are finite and cannot cover the continuous state space. If the testing distribution and training distribution are the same, one can leverage the Rademacher complexity to give an error bound that the CBF conditions are violated, as is in Appendix~B of~\cite{qin2021learning}. But if the testing distribution is different from training, it is still unclear to derive the generalization error of the CBF conditions. To train CBF and controller that provably satisfy the CBF conditions, one can also use verification tools to find the counterexamples in the state space that violates the CBF conditions and add those counterexamples to the training set~\cite{dai2021lyapunov, chang2019neural}. The process is finished when no more counterexample can be found. However, the time complexity of the verification makes it not applicable for large and expressive neural networks. Also, the error between the nominal and real dynamics will have a negative impact on the safety performance. These limitations are left for future work.

\section{Conclusion and future works}
We presented SABLAS, a general-purpose safe controller learning approach for black-box systems, which is supported by the theoretical guarantees from control barrier function theory, and at the same time is strengthened using a novel learning structure so it can directly learn the policies and barrier certificates for black-box dynamical systems. 
Simulation results show that SABLAS indeed provides a systematic way of learning safe control policies with a great improvement over safe RL methods. For future works, we plan to study SABLAS on multi-agent systems, especially with adversarial players. 

\bibliographystyle{IEEEtran}
\bibliography{IEEEabrv,references}

\end{document}